\documentclass[]{spie}  

\usepackage{amsmath,amsfonts,amssymb}
\usepackage{graphicx}
\usepackage[colorlinks=true, allcolors=blue]{hyperref}

\usepackage{times}
\usepackage{float}
\usepackage{booktabs}
\usepackage{epstopdf}
\usepackage[table]{xcolor}
\usepackage{amsmath}
\usepackage{numprint}
\usepackage{subcaption}
\usepackage{caption}
\usepackage{multirow}
\usepackage{array}
\usepackage{calc}
\usepackage{calc}
\usepackage{tikz}
\usetikzlibrary{positioning}
\usepackage{algorithmic}
\usepackage{algorithm}

\title{Deep Learning based Defect classification and detection in SEM images: A Mask R-CNN approach}

\author[a,c]{Bappaditya Dey}
\author[a,f]{Enrique Dehaerne}
\author[b,d]{Kasem Khalil}
\author[a]{Sandip Halder}
\author[a]{Philippe Leray}
\author[c,e]{Magdy A. Bayoumi}
\affil[a]{imec, Kapeldreef 75, 3001 Leuven, Belgium}
\affil[b]{Department of Electrical and Computer Engineering, University of Mississippi, Mississippi, USA}
\affil[c]{The Center for Advanced Computer Studies,  University of Louisiana at Lafayette, Louisiana, USA}
\affil[d]{Department of Electrical Engineering, Assiut University, Egypt}
\affil[e] {Department of Electrical and Computer Engineering, University of Louisiana at Lafayette, Louisiana, USA}
\affil[f]{Department of Computer Science, KU Leuven, Leuven, Belgium}

\authorinfo{Further author information: (Send correspondence to Bappaditya Dey and Kasem Khalil)\\Bappaditya Dey: E-mail: Bappaditya.Dey@imec.be}

\begin{document} 
\maketitle

\begin{abstract}
	In this research work, we have demonstrated the application of Mask-RCNN (Regional Convolutional Neural Network), a deep-learning algorithm for computer vision and specifically object detection, to semiconductor defect inspection domain. Stochastic defect detection and classification during semiconductor manufacturing has grown to be a challenging task as we continuously shrink circuit pattern dimensions (e.g., for pitches less than 32 nm).
	Defect inspection and analysis by state-of-the-art optical and e-beam inspection tools is generally driven by some rule-based techniques, which in turn often causes to misclassification and thereby necessitating human expert intervention. In this work, we have revisited and extended our previous deep learning-based defect classification and detection method towards improved defect instance segmentation in SEM images with precise extent of defect as well as generating a mask for each defect category/instance. This also enables to extract and calibrate each segmented mask and quantify the pixels that make up each mask, which in turn enables us to count each categorical defect instances as well as to calculate the surface area in terms of pixels. We are aiming at detecting and segmenting different types of inter-class stochastic defect patterns such as bridge, break, and line collapse as well as to differentiate accurately between intra-class multi-categorical defect bridge scenarios (as thin/single/multi-line/horizontal/non-horizontal) for aggressive pitches as well as thin resists (High NA applications). Our proposed approach demonstrates its effectiveness both quantitatively and qualitatively.\\
\end{abstract}



\section{Introduction}
\label{sect:intro}  
As we continuously shrink device dimensions from node to node to make chips smaller in form-factor, numerous challenges appear in different areas of patterning. Until now scaling has been primarily possible due to wavelength scaling on the scanner side, i.e., by continuously moving to lower and lower wavelengths, thereby making smaller pitches to be directly printed possible and by design and integration tricks. Currently, the smallest wavelength, scanners use, is 13.5nm (EUV) and this is being used for manufacturing advanced logic and memory devices.  While on one side we manage to make the devices smaller, on the other side it brings in new challenges in the manufacturing and yield domain. 


\begin{figure}[!tb]
	\begin{subfigure}{0.35\textwidth}
		\centering
		\includegraphics[width=0.80\linewidth]{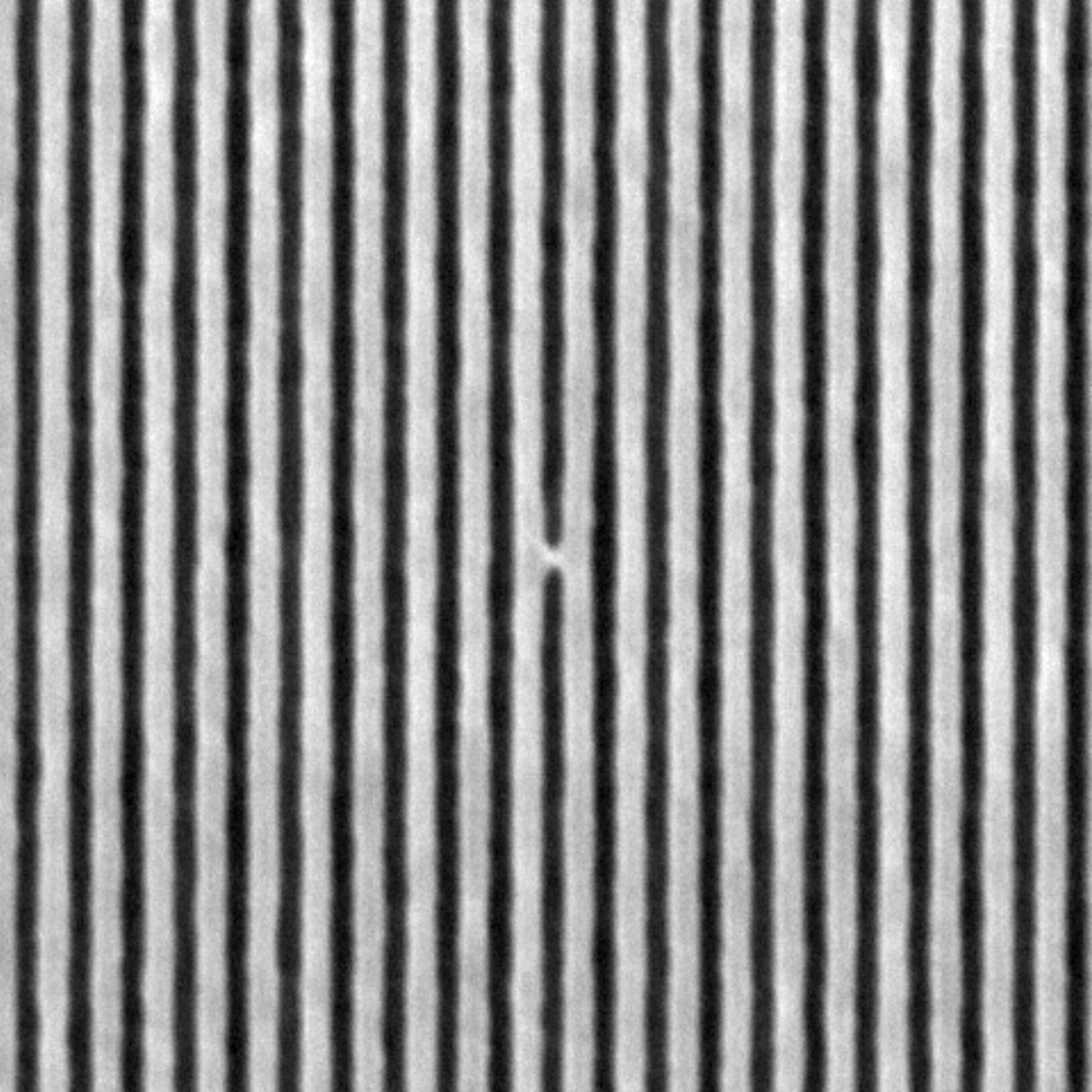}
		\captionsetup{singlelinecheck=off}
		\caption[.]{\label{fig:fig_1a}}
	\end{subfigure}
	~
	\begin{subfigure}{0.35\textwidth}
		\centering
		\includegraphics[width=0.80\linewidth]{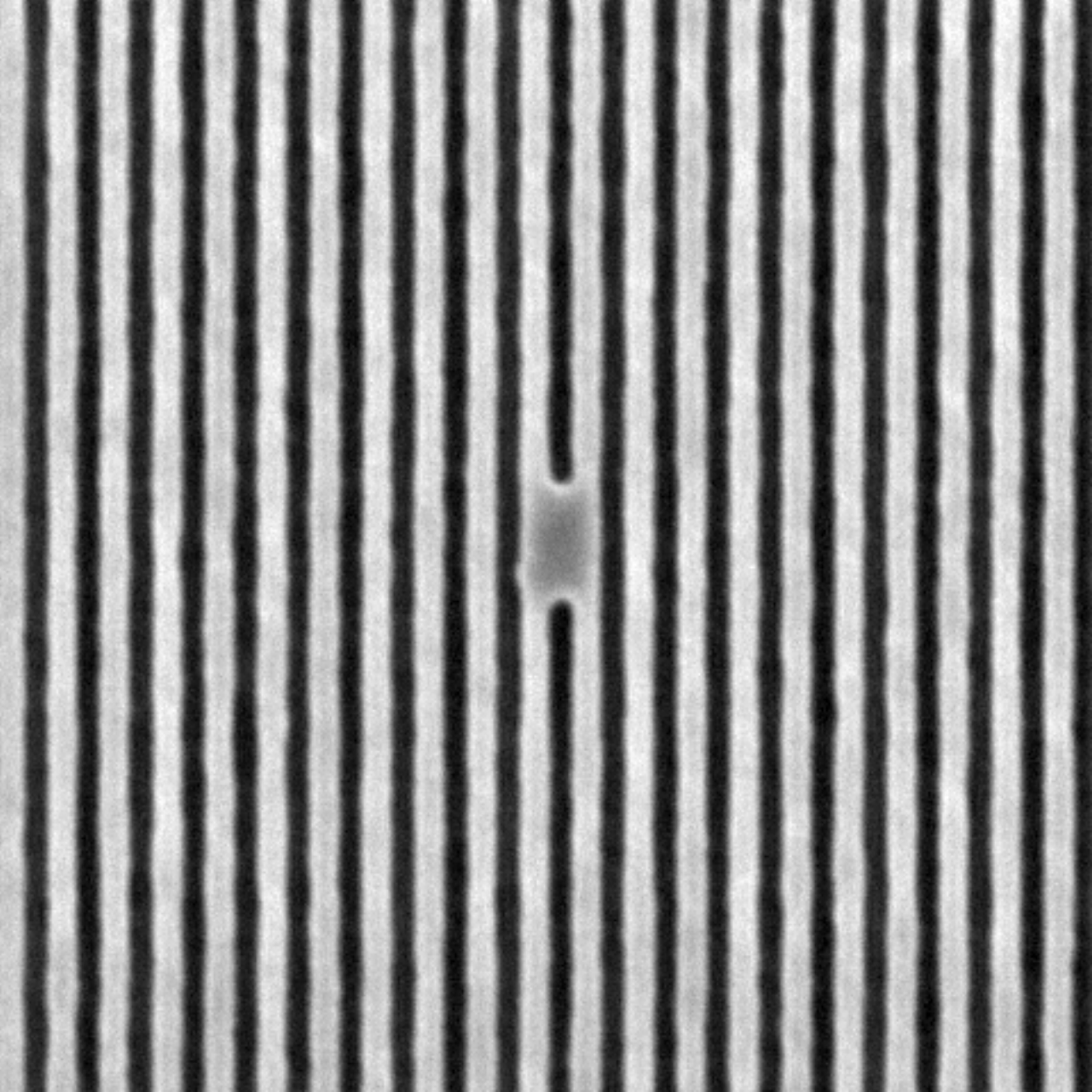}
		\captionsetup{singlelinecheck=off}
		\caption[.]{\label{fig:fig_1b}}
	\end{subfigure}
~
	\begin{subfigure}{0.35\textwidth}
		\centering
		\includegraphics[width=0.80\linewidth]{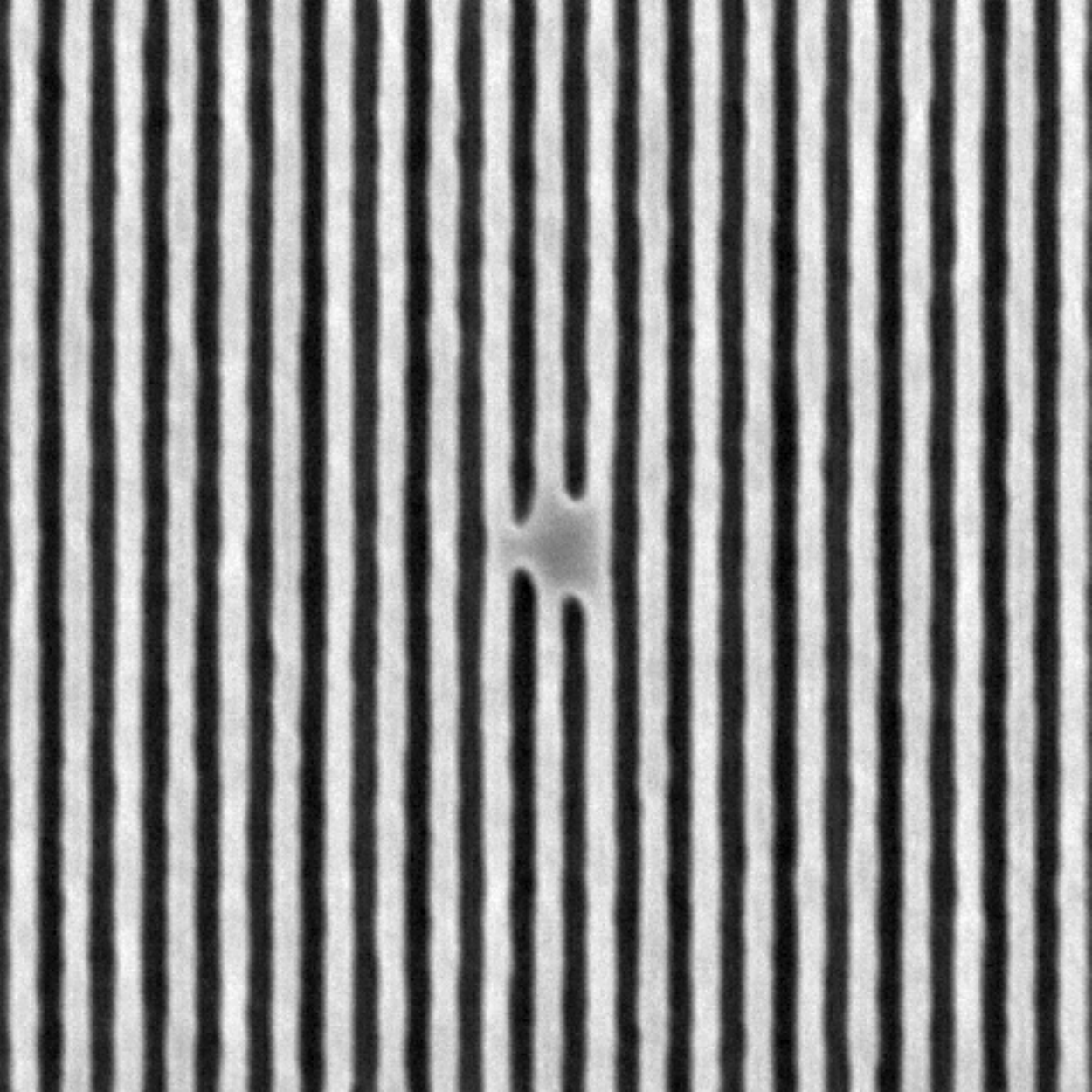}
		\captionsetup{singlelinecheck=off}
		\caption[.]{\label{fig:fig_1c}}
	\end{subfigure}
	\\
	\begin{subfigure}{0.35\textwidth}
		\centering
		\includegraphics[width=0.80\linewidth]{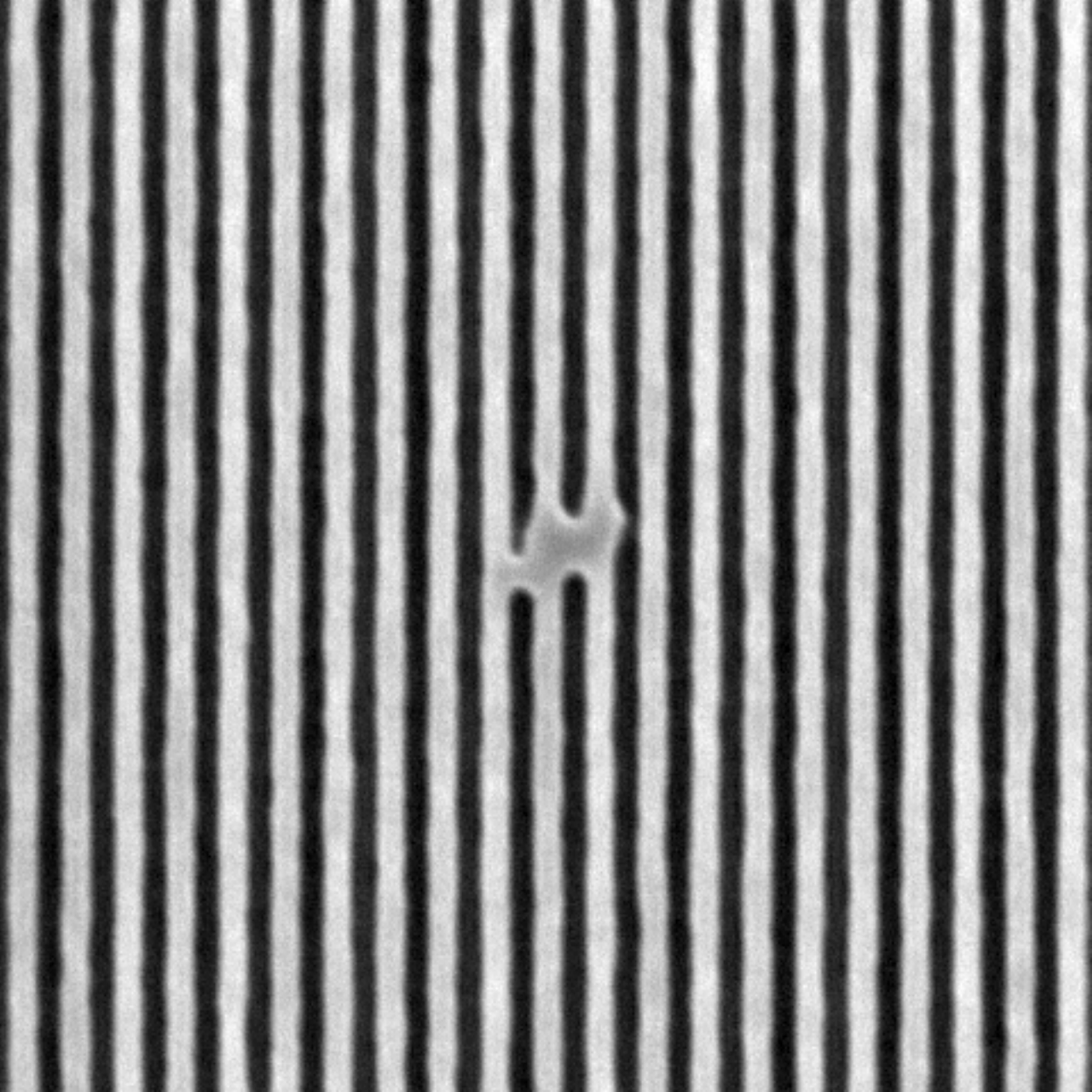}
		\captionsetup{singlelinecheck=off}
		\caption[.]{\label{fig:fig_1d}}
	\end{subfigure}
~
	\begin{subfigure}{0.35\textwidth}
		\centering
		\includegraphics[width=0.80\linewidth]{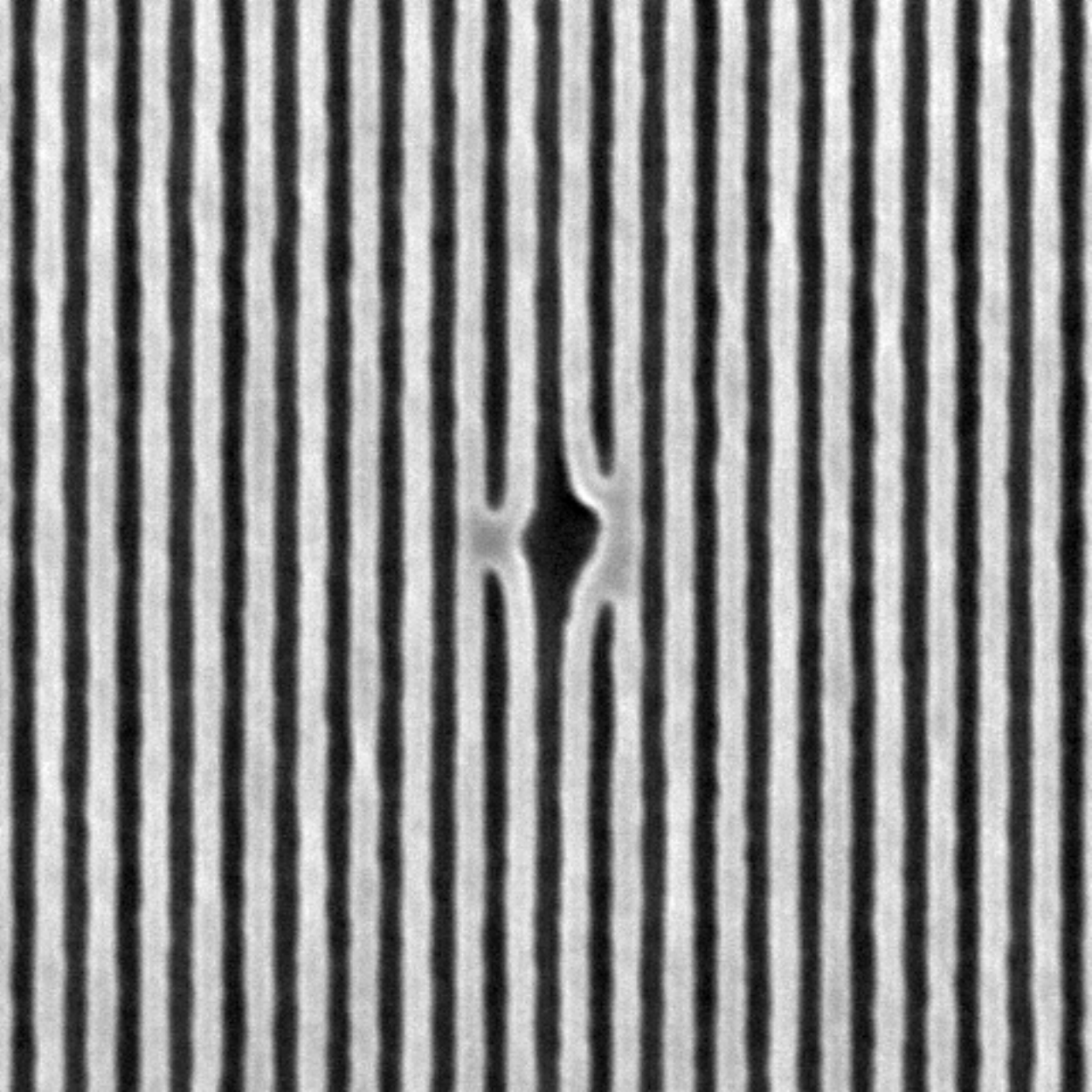}
		\captionsetup{singlelinecheck=off}
		\caption[.]{\label{fig:fig_1e}}
	\end{subfigure}
	~
	\begin{subfigure}{0.35\textwidth}
		\centering
		\includegraphics[width=0.80\linewidth]{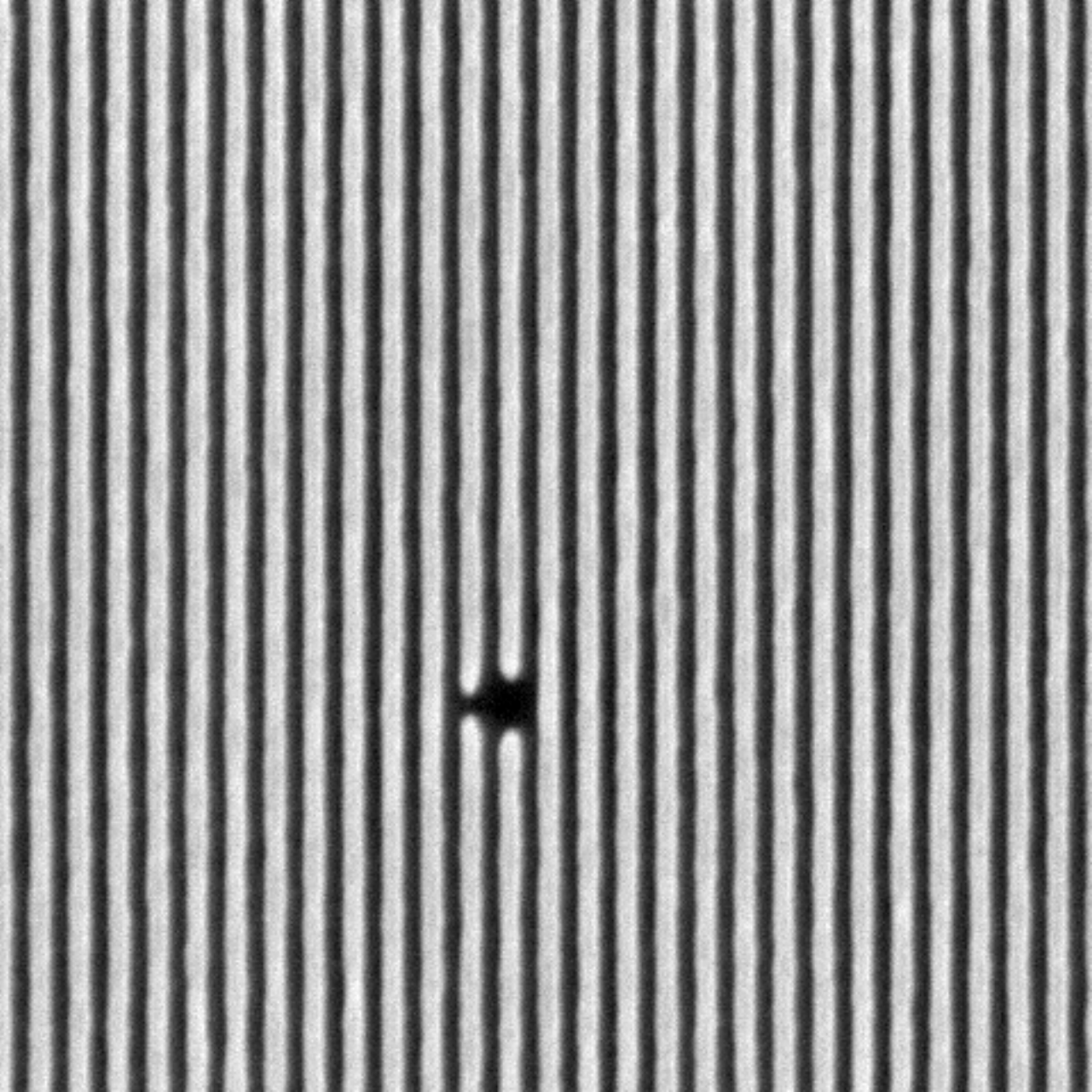}
		\captionsetup{singlelinecheck=off}
		\caption[.]{\label{fig:fig_1f}}
	\end{subfigure}
	
	\captionsetup{singlelinecheck=off}
	\caption[.]{Typical Defects: (a) Thin-Bridge, (b) Single-Bridge, (c) Multi-Line-Bridge-Horizontal, (d) Multi-Line-Bridge-Non-Horizontal, (e) Line-Collapse, and (f) Line-Break.\label{fig:fig1}}
\end{figure}


Metrology and inspection of devices become more and more challenging as device dimensions become smaller and smaller. Detection and prevention of these small defects are of utmost importance for a product to yield. Optical tools have been the workhorse of the fabs for detecting, preventing, and understanding all sorts of defects. However, as we shrink, to make patterns with pitches below 32 nm, sometimes the sub 10nm defects are missed by the optical tools. The roughness of the lines and patterns often increase the noise floor of the measurement thereby complicating the detection of these small defects. New and advanced algorithms on image processing are continuously being applied on the optical tools to boost the signal from these extremely miniscule defects. In certain cases, e-beam based tools are quickly becoming the preferred alternative option. However, e-beam tools also have their own challenges. First and foremost, it is immensely slow when compared to state-of-the-art optical inspection tools. E-beam tools can be used for inspection/review but only on small areas/predetermined locations. Also defect classification of the SEM images remains a major challenge. Furthermore, because high resolution images are gathered from the SEM images, yield engineers want better classification of the different sorts of defects. Defect classification becomes complicated as we have more and more classes.

%
%

Fig.~\ref{fig:fig1} shows SEM images with examples of different defect categories generally encountered in aggressive pitches. Fig.~\ref{fig:fig1}[(a)-(f)] are examples of Line-Space (L/S) patterns with defect types as (a) Thin-Bridge, (b) Single-Bridge, (c) Multi-Line-Bridge-Horizontal, (d) Multi-Line-Bridge-Non-Horizontal, (e) Line-Collapse, and (f) Line-Break, respectively. These stochastic defect patterns may also vary randomly with variable degrees of pixel-level extent, both for inter-class as well as intra-class level.

In this research, we have proposed an Automated Defect Classification and Detection (ADCD) framework for improved defect instance segmentation in SEM images with precise extent of defect as well as generating a mask for each defect category/instance. In our previous work Refs.~\cite{dey2222, Dey2022, Online3}, we have demonstrated a novel robust supervised Deep Learning (DL) training scheme to detect different defects from SEM images and accurately classify them according to their corresponding classes in aggressive pitches as well as thin resists (High NA applications). This work enables to extract and calibrate each segmented mask and quantify the pixels that make up each mask, which in turn enables us to count each categorical defect instances as well as to calculate the surface area in terms of pixels. Finally, we have proposed an Automated Defect Classification-Detection-Segmentation (ADCDS) framework using state-of-the-art Mask R-CNN Ref.~\cite{he2017mask} model to detect and segment different types of inter-class stochastic defect patterns such as bridge, break, and line collapse as well as to differentiate accurately between intra-class multi-categorical defect bridge scenarios (as thin/single/multi-line/horizontal/non-horizontal) for aggressive pitches as well as thin resists (High NA applications).

The remainder of the paper is organized as follows. Sec. 2 introduces some related work. In Sec. 3, we provide an overview of the Mask R-CNN architecture backbone and our proposed method. Sec. 4 demonstrates the experiments done followed by Sec. 5 reporting on the experimental results and analysis. In Sec. 6, we conclude the paper.


\section{Related Work} 
\label{sect:Related Work}

In this section we briefly introduce related Machine Learning (ML)-based semiconductor defect detection literature. Ref.~\cite{lopez2021review} reviewed nine ML-based semiconductor SEM defect detection studies published in and before 2020. They found that Convolutional neural networks (CNNs) Ref.~\cite{khalil2022designing} outperform other ML methods such as Support Vector Machines (SVMs) or random forests. A caveat is that the reviewed CNN models required relatively large training sets to achieve these results. It was suggested by Ref.~\cite{phua2022dladc} that non-deep learning approaches give poor results due to the need for manual feature selection which is prone to error. They propose a CNN model that detects and classifies defects in SEM images and subcategorizes these defects based on size which can help identify the root cause of the defect. Ref.~\cite{kim2021adversarial} uses a conditional Generative Adversarial Networks (GANs) with convolutional layers to detect defects. The generator outputs the bounding boxes and severity of the defects (“hard” or “soft”) given pairs of images and layouts. The discriminator is conditioned on the predicted defect classes as well as the pairs of images and layouts. An ensemble of CNN models was implemented by Ref.~\cite{Online3} to detect and classify defects of various types (e.g. “gap”, “bridge”, and “line collapse”). Different models of varying sizes were found to perform better on different defect classes which allows the ensemble of these models to perform best in the general case. To the best of our knowledge, no previous work has been conducted on semiconductor defect segmentation.

\begin{figure}
	\begin{center}
		\begin{tabular}{c}
			\includegraphics[width=1.0\linewidth]{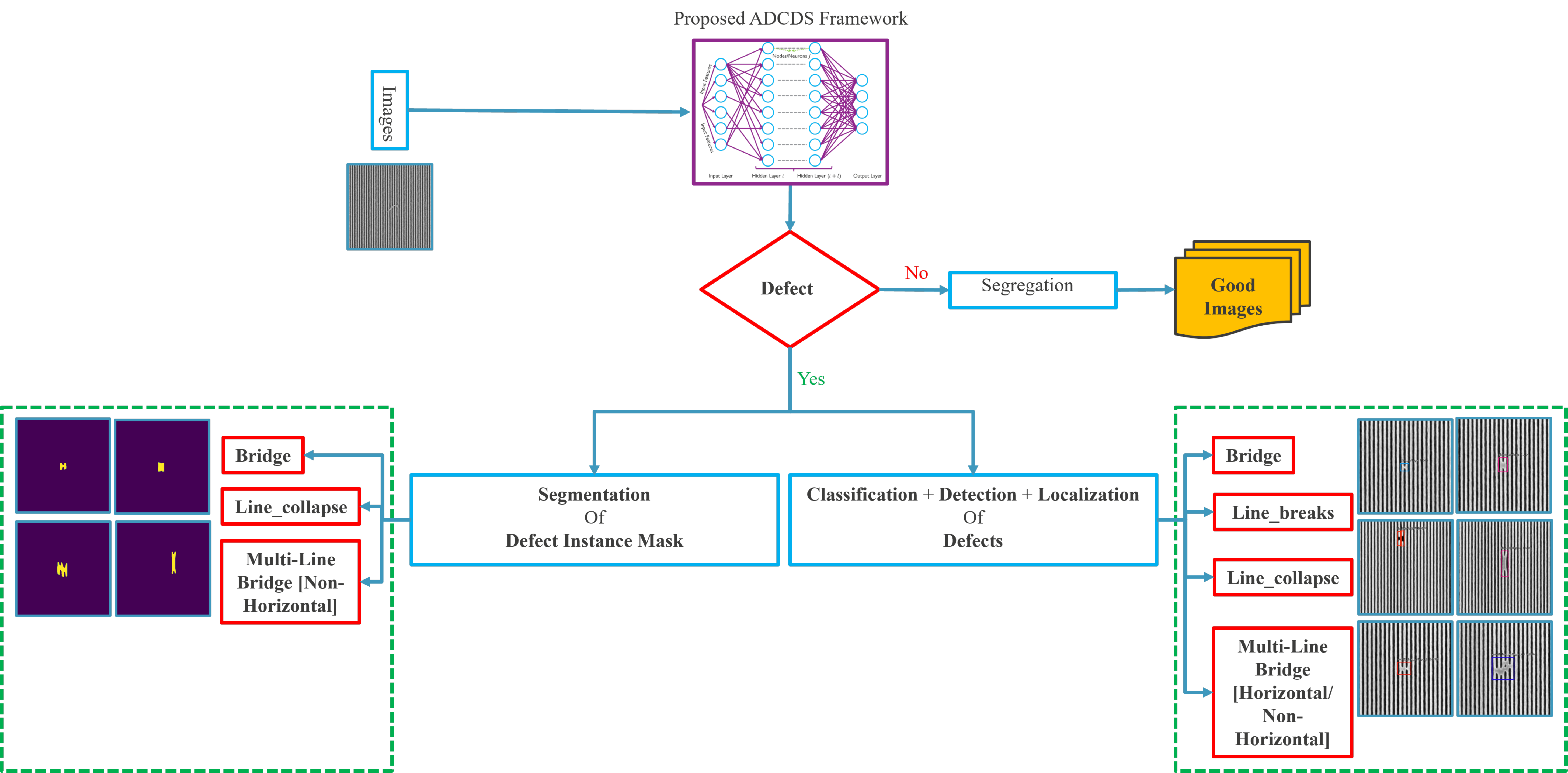}
		\end{tabular}
	\end{center}
	\caption 
	{ \label{fig:fig_2}
		Proposed Mask R-CNN model-based ADCDS framework. } 
\end{figure} 


\begin{figure}
	\begin{center}
		\begin{tabular}{c}
			\includegraphics[width=1.0\linewidth]{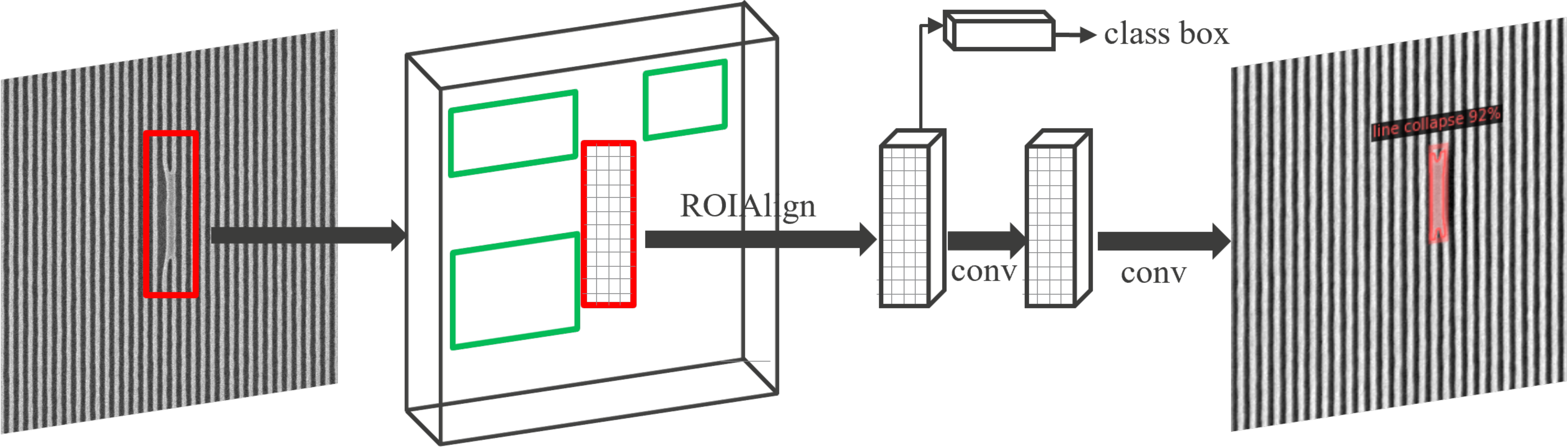}
		\end{tabular}
	\end{center}
	\caption 
	{ \label{fig:fig_3}
		Mask R-CNN architecture for defect instance segmentation. } 
\end{figure} 


\begin{figure}
	\begin{center}
		\begin{tabular}{c}
			\includegraphics[width=0.50\linewidth]{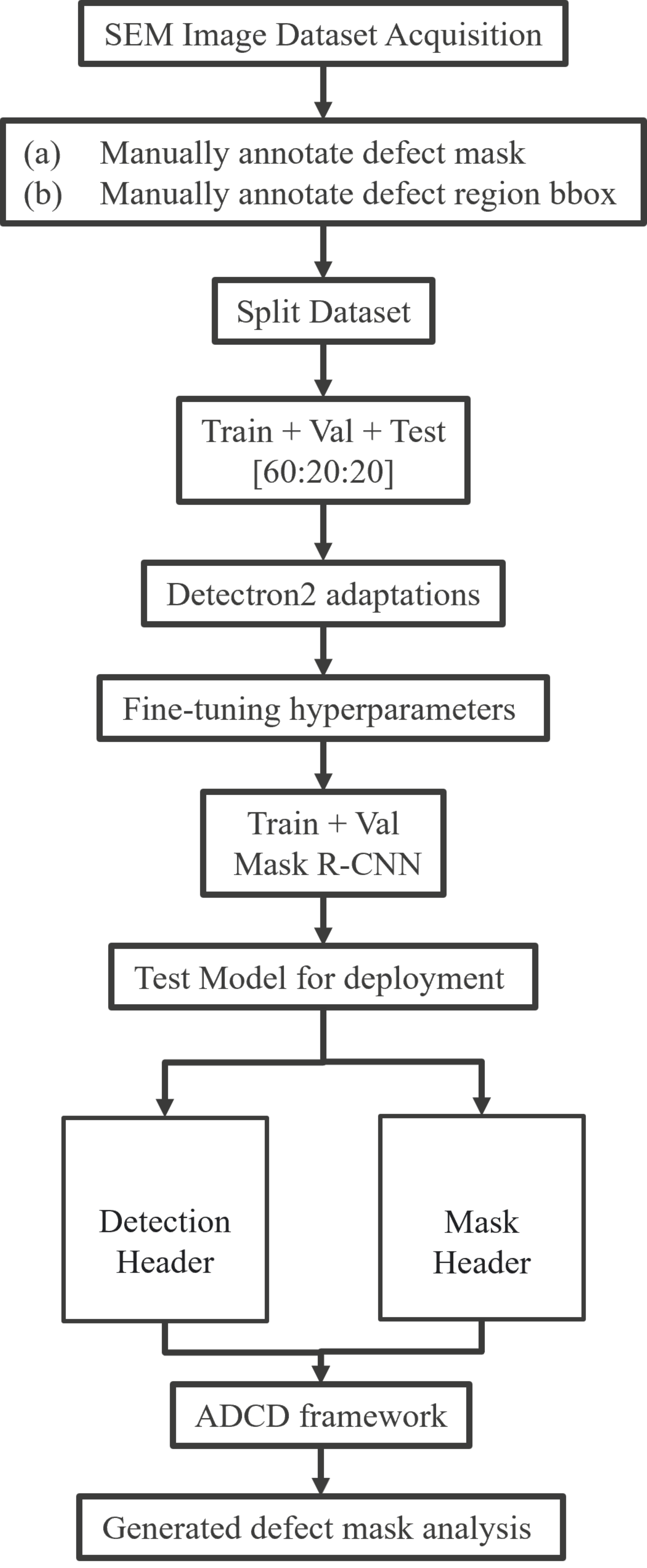}
		\end{tabular}
	\end{center}
	\caption 
	{ \label{fig:fig_4}
		Flowchart of Mask R-CNN based ADCDS framework deployment. } 
\end{figure} 



\begin{figure}
	\centering
	\begin{subfigure}{0.35\textwidth}
		\centering
		\includegraphics[width=0.80\linewidth]{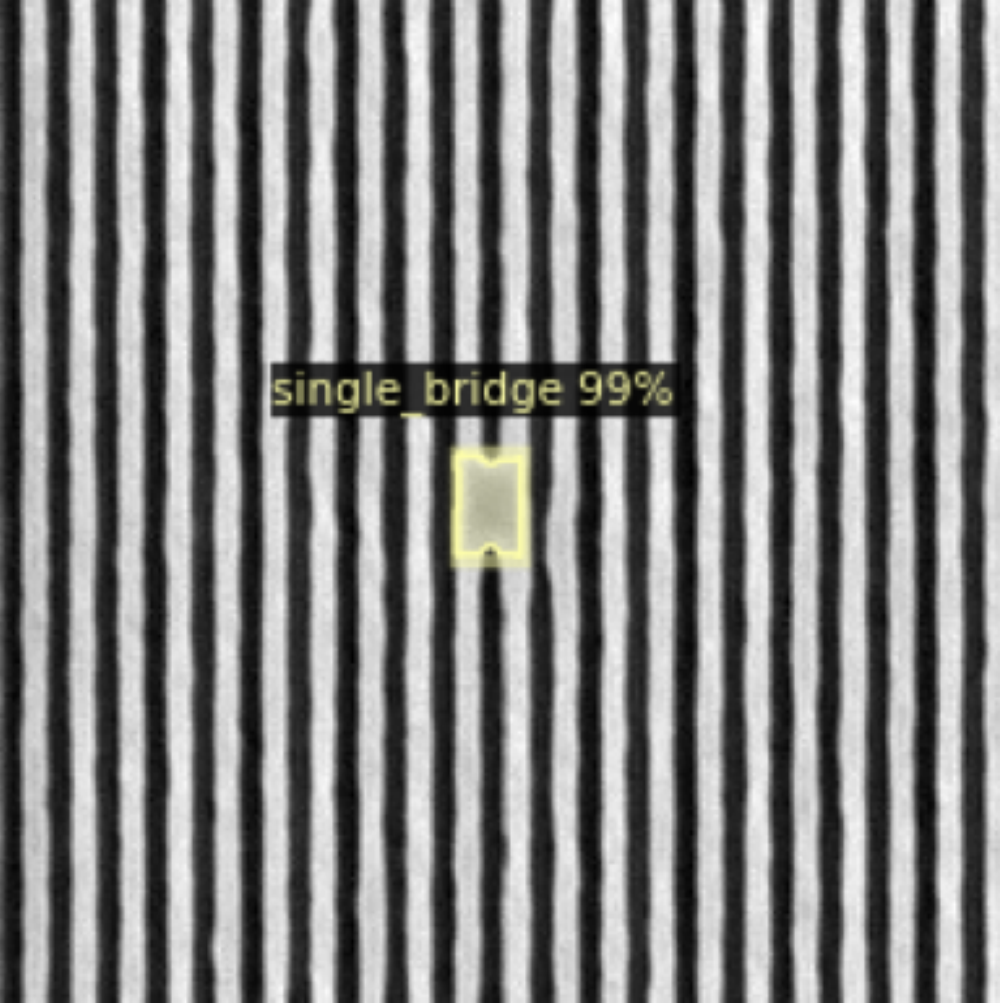}
		\captionsetup{singlelinecheck=off}
		\caption[.]{\label{fig:fig_5a}}
	\end{subfigure}
	~
	\begin{subfigure}{0.35\textwidth}
		\centering
		\includegraphics[width=0.80\linewidth]{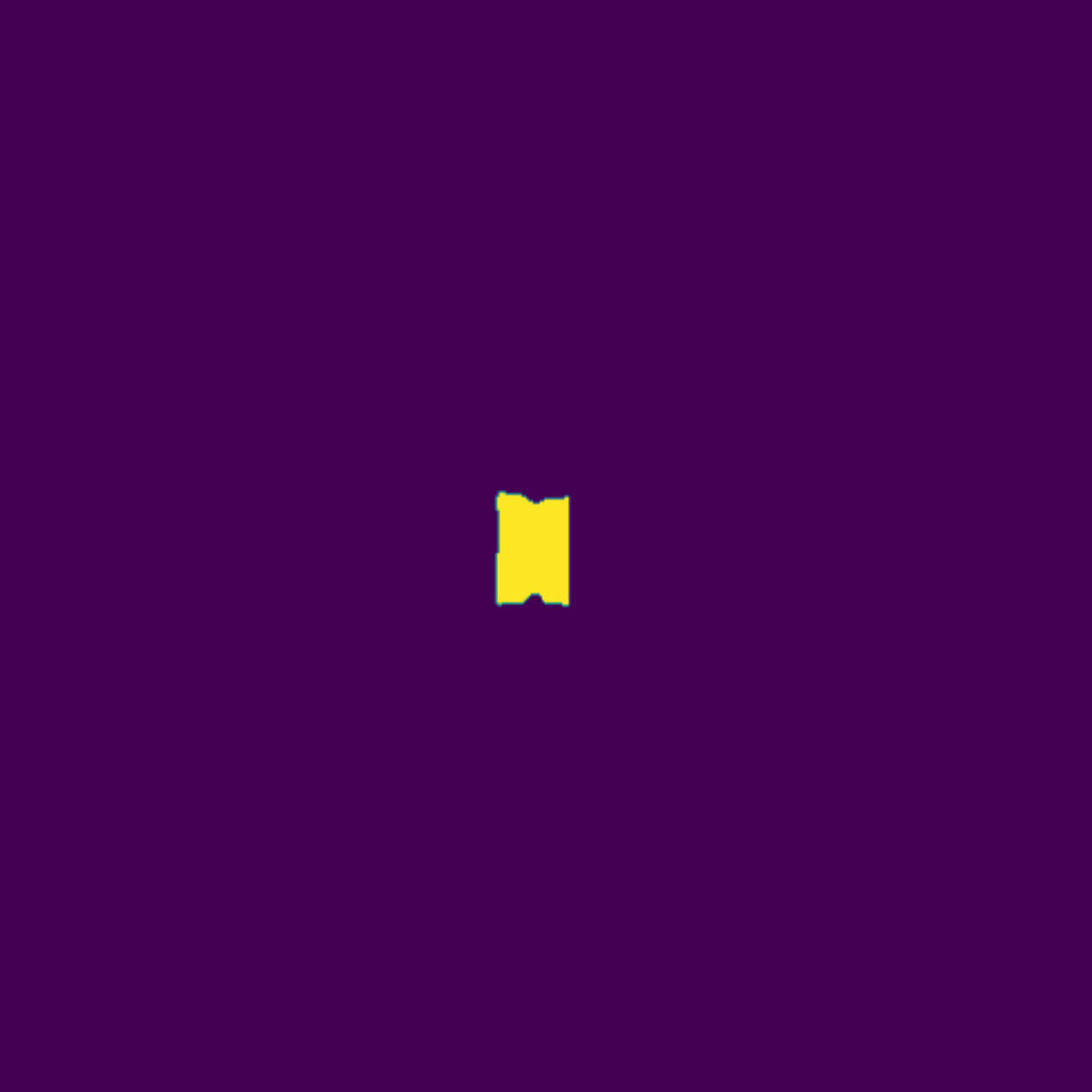}
		\captionsetup{singlelinecheck=off}
		\caption[.]{\label{fig:fig_5b}}
	\end{subfigure}
	\\
	\begin{subfigure}{0.35\textwidth}
		\centering
		\includegraphics[width=0.80\linewidth]{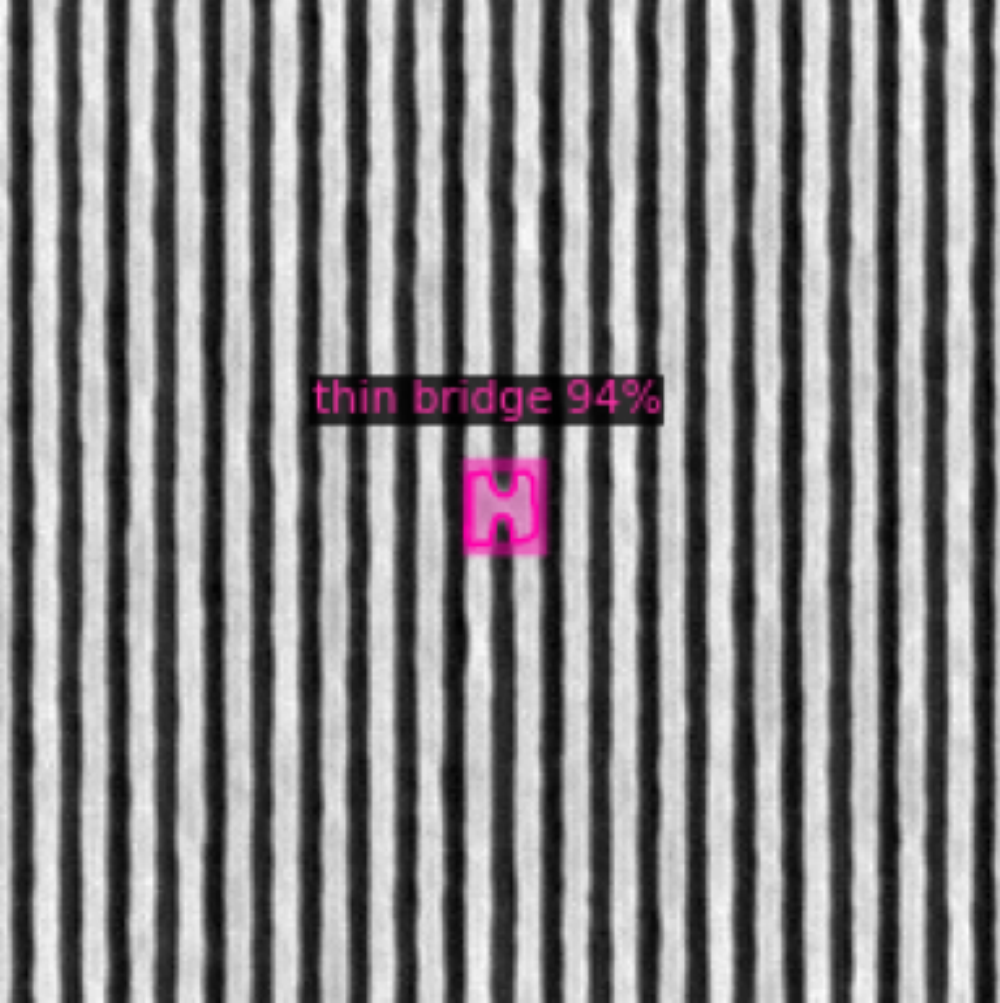}
		\captionsetup{singlelinecheck=off}
		\caption[.]{\label{fig:fig_5c}}
	\end{subfigure}
	~
	\begin{subfigure}{0.35\textwidth}
		\centering
		\includegraphics[width=0.80\linewidth]{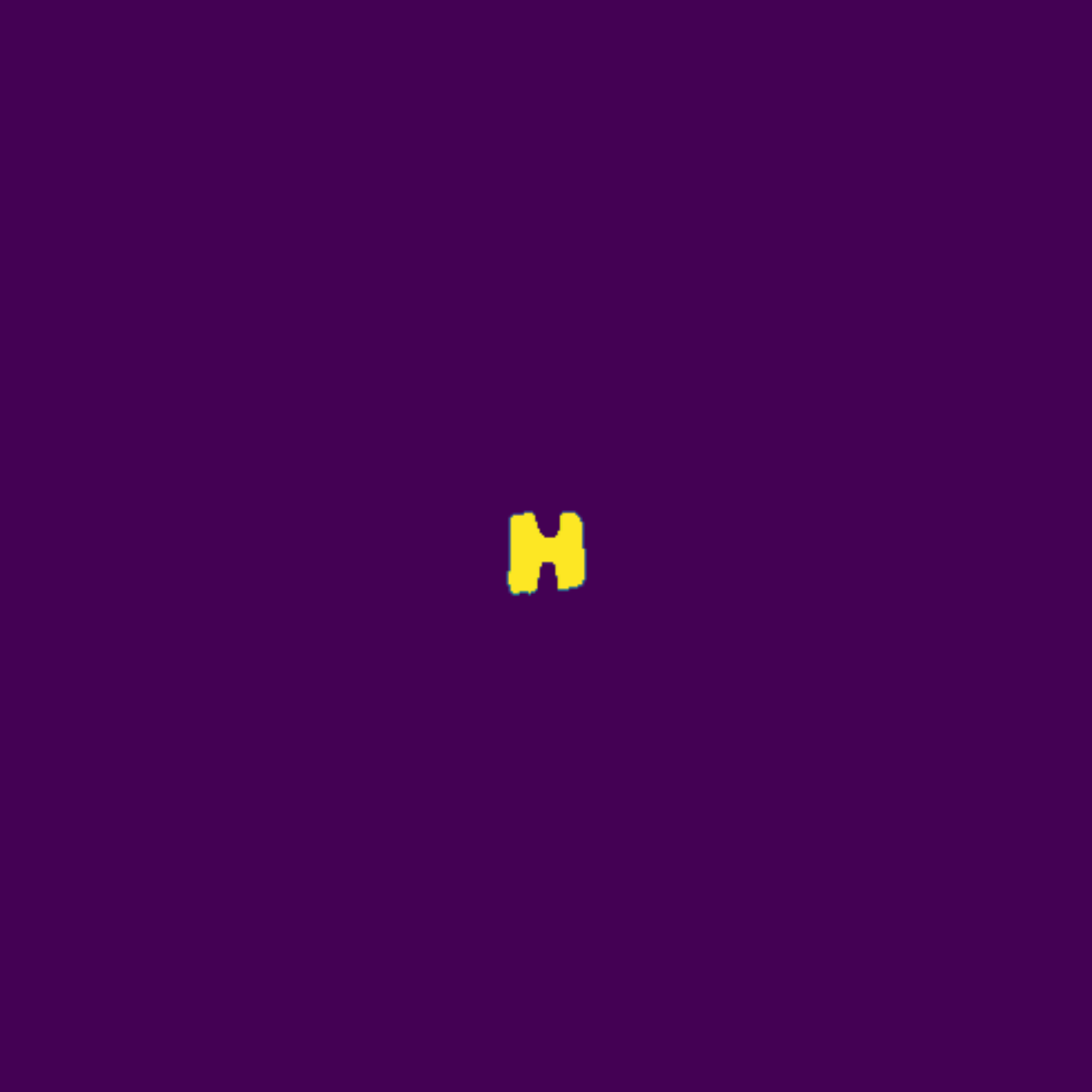}
		\captionsetup{singlelinecheck=off}
		\caption[.]{\label{fig:fig_5d}}
	\end{subfigure}
	\\
	\begin{subfigure}{0.35\textwidth}
		\centering
		\includegraphics[width=0.80\linewidth]{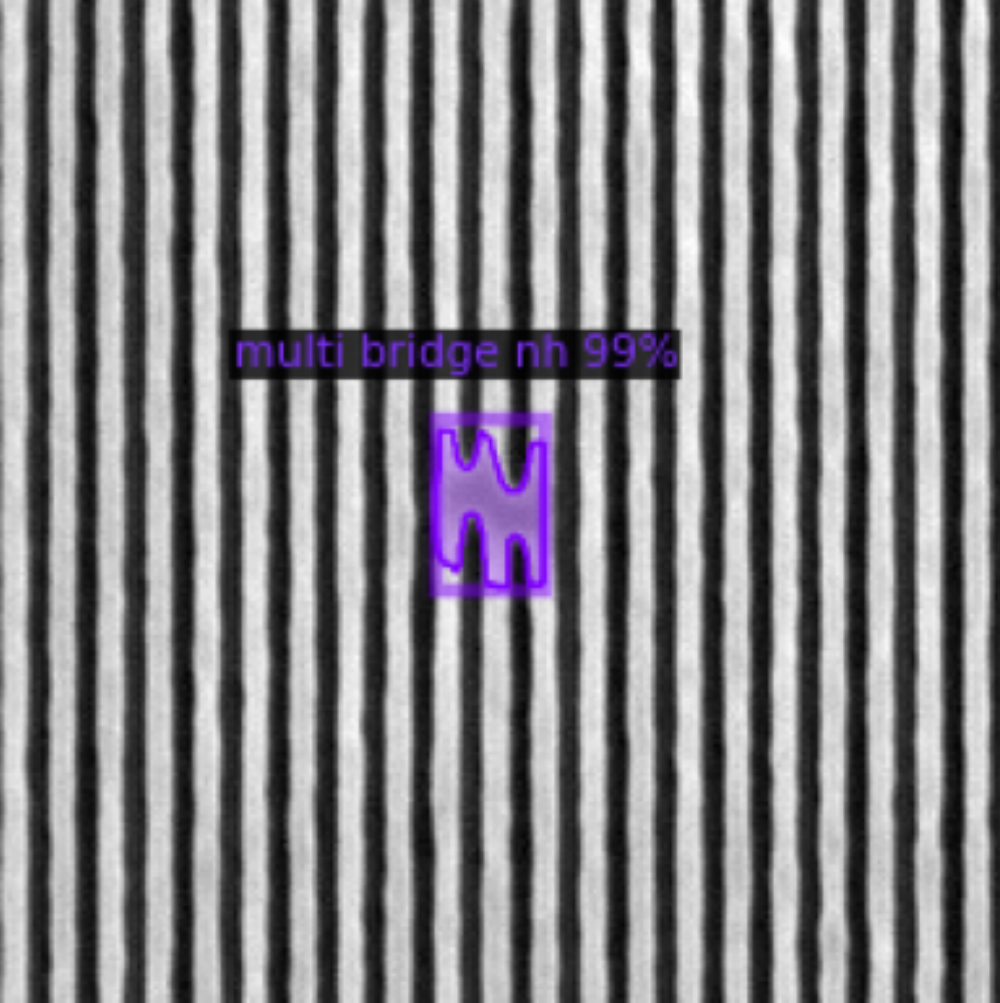}
		\captionsetup{singlelinecheck=off}
		\caption[.]{\label{fig:fig_5e}}
	\end{subfigure}
	~
	\begin{subfigure}{0.35\textwidth}
		\centering
		\includegraphics[width=0.80\linewidth]{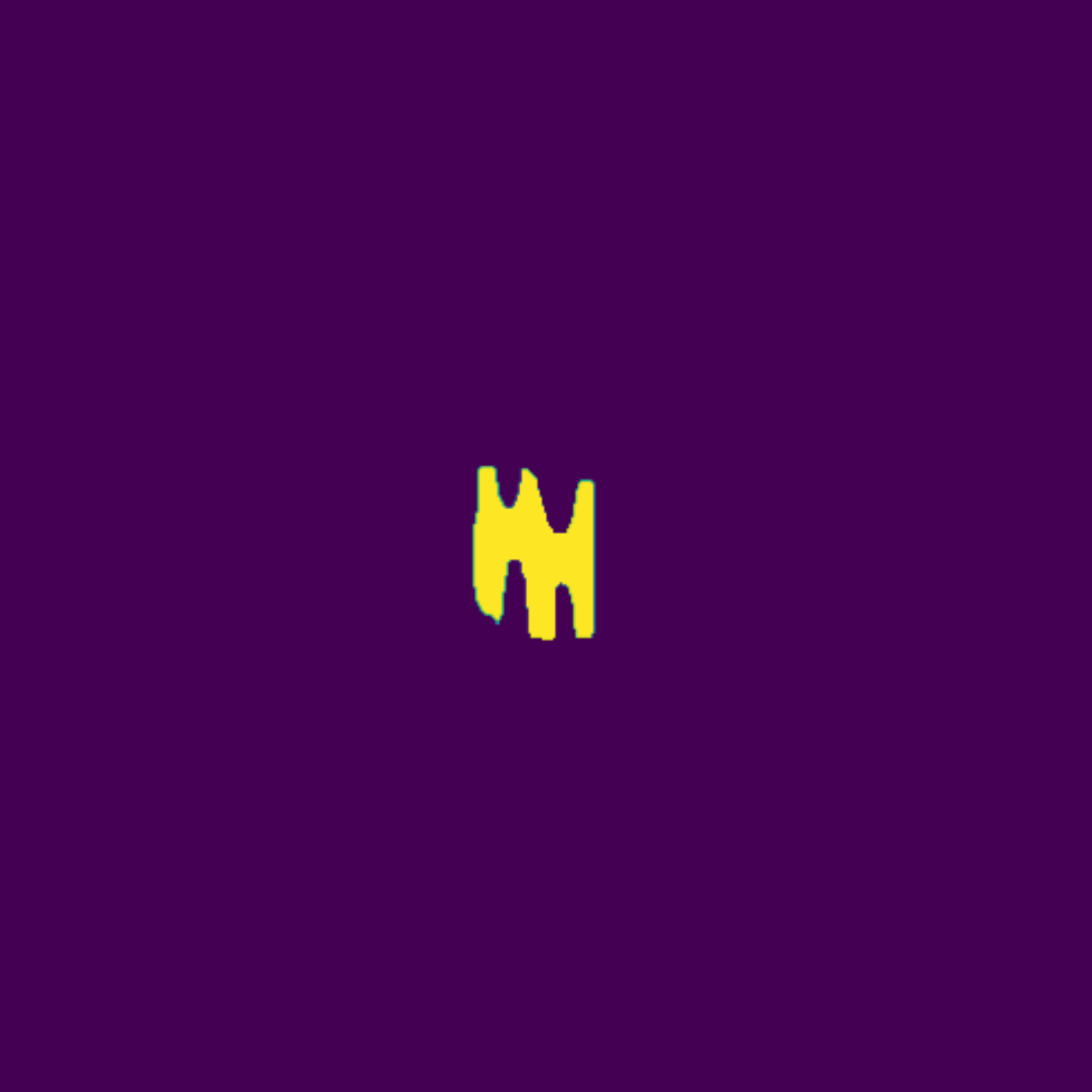}
		\captionsetup{singlelinecheck=off}
		\caption[.]{\label{fig:fig_5f}}
	\end{subfigure}
	\\
	\begin{subfigure}{0.35\textwidth}
		\centering
		\includegraphics[width=0.80\linewidth]{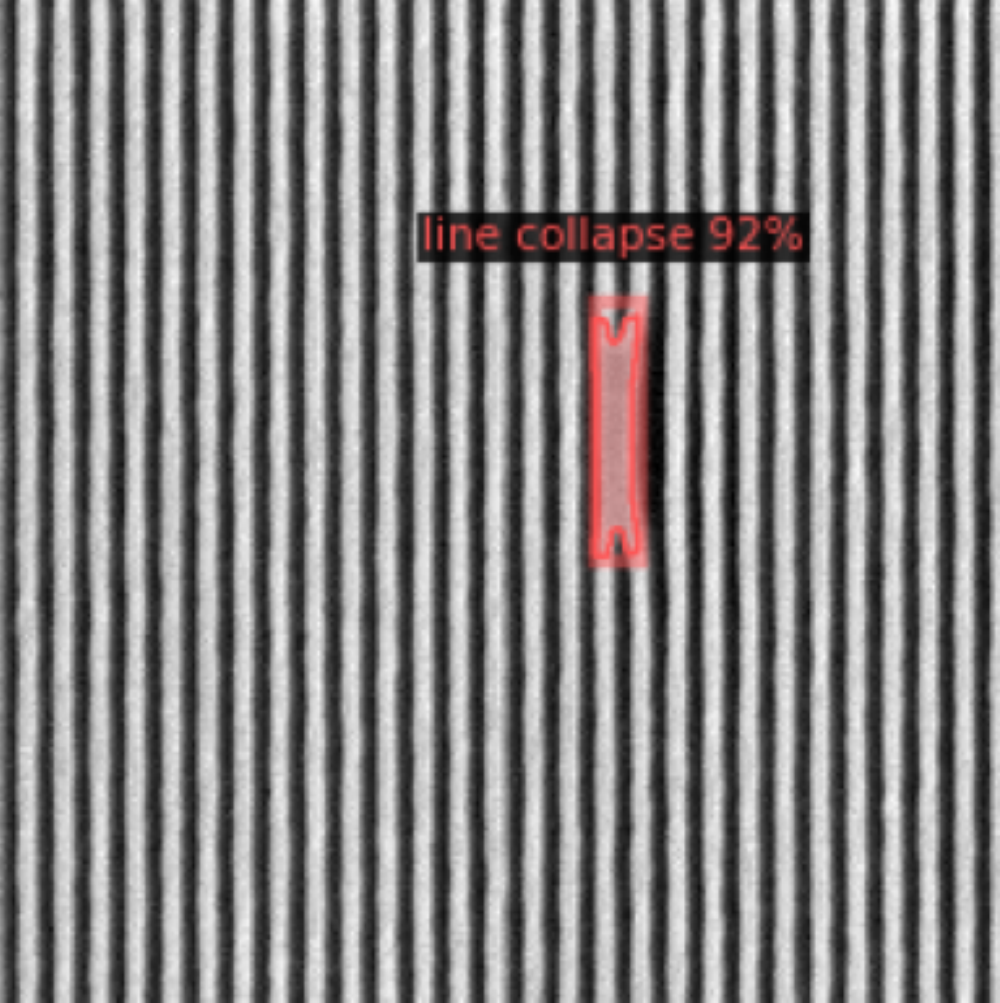}
		\captionsetup{singlelinecheck=off}
		\caption[.]{\label{fig:fig_5g}}
	\end{subfigure}
	~
	\begin{subfigure}{0.35\textwidth}
		\centering
		\includegraphics[width=0.80\linewidth]{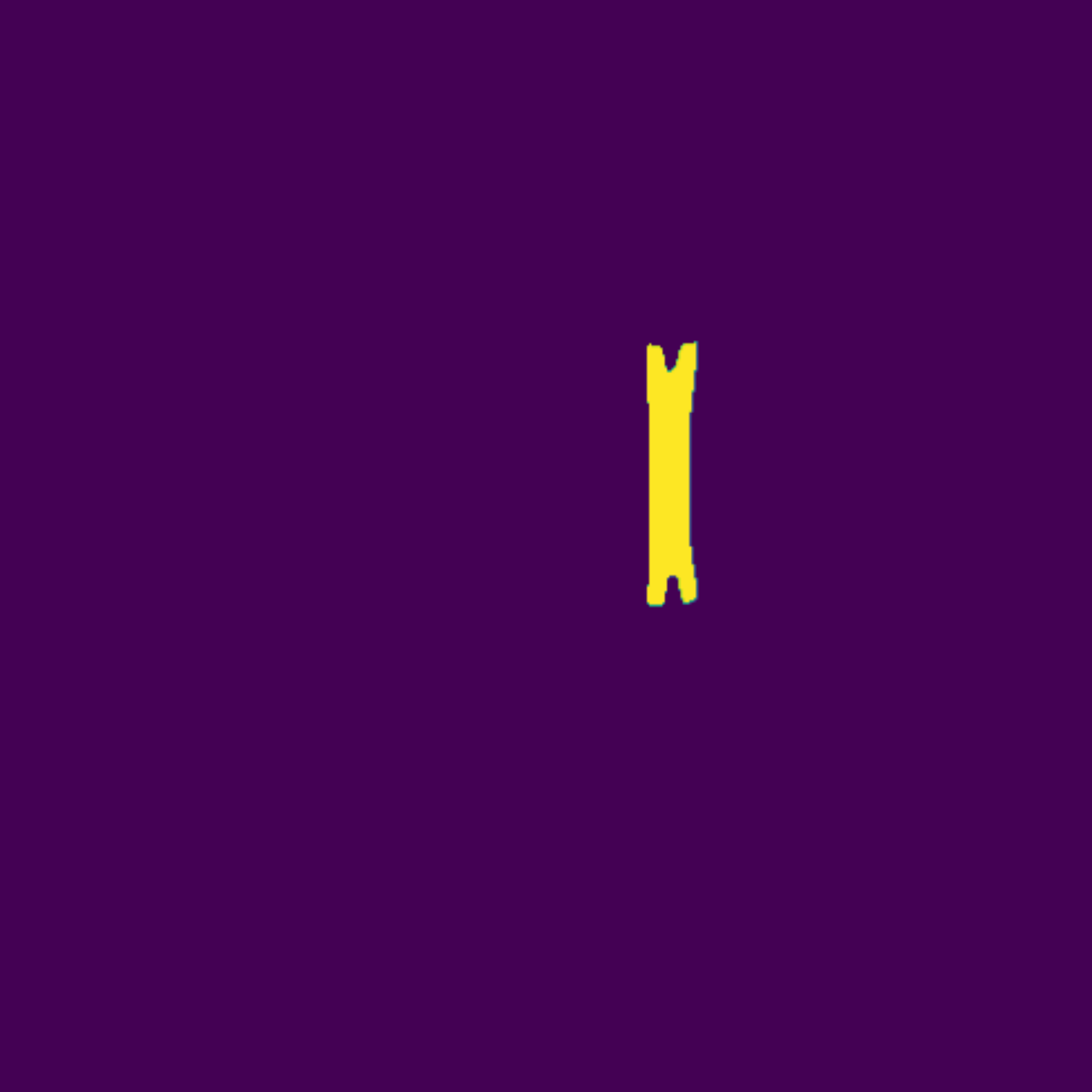}
		\captionsetup{singlelinecheck=off}
		\caption[.]{\label{fig:fig_5h}}
	\end{subfigure}

	\captionsetup{singlelinecheck=off}
	\caption[.]{Defect detection results with proposed Mask R-CNN approach. (a) Single bridge, (c) Thin bridge, (e) Multi-line non-horizontal bridge, and (g) Line collapse.
		Output masks for corresponding (b) Single bridge, (d) Thin bridge, (f) Multi-line non-horizontal bridge, and (h) Line collapse defects.\label{fig:fig5}}
\end{figure}


\begin{figure} [!tb]
	\begin{center}
		\begin{tabular}{c}
			\includegraphics[width=0.90\linewidth]{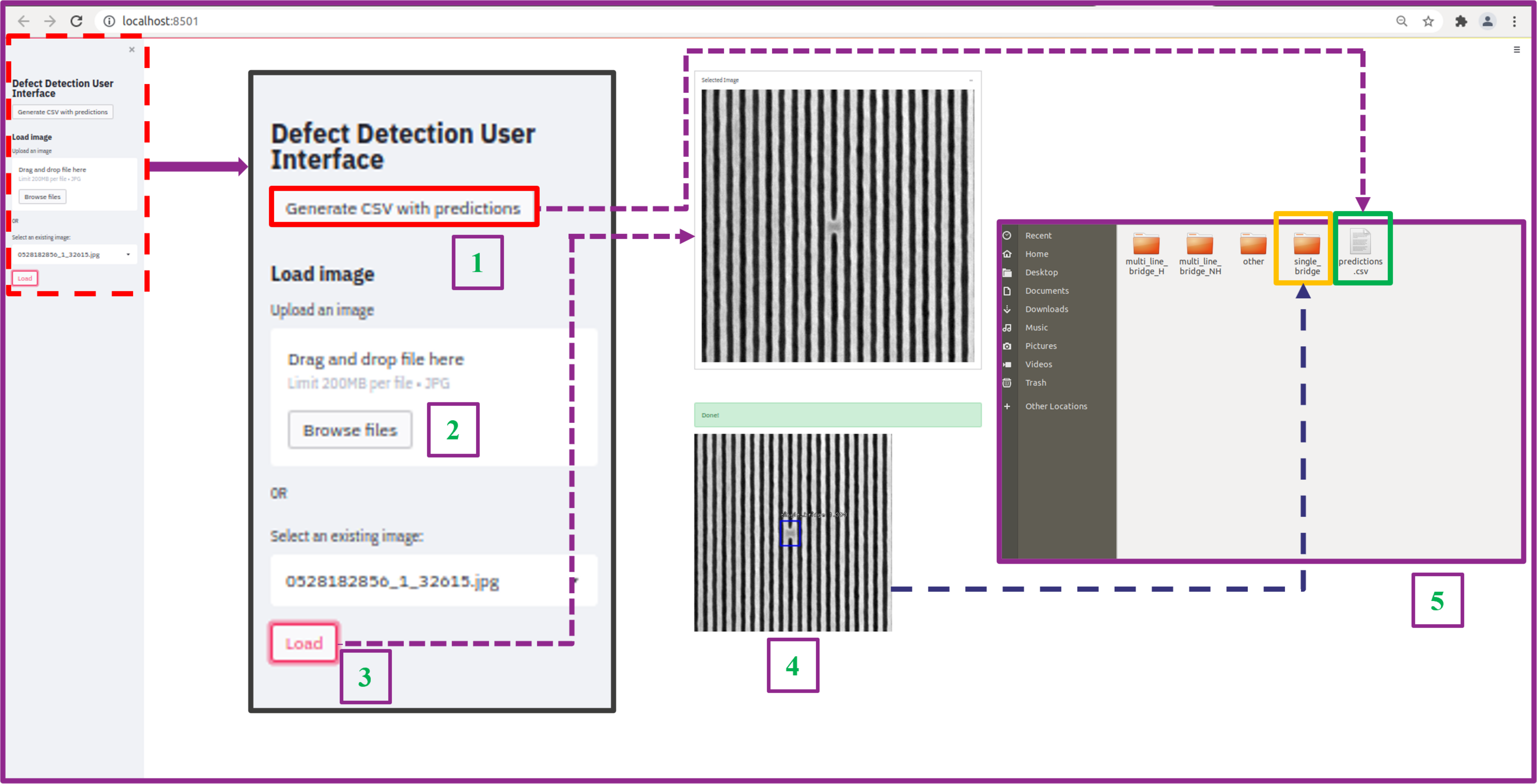}
		\end{tabular}
	\end{center}
	\caption 
	{ \label{fig:fig_6}
		Web-based defect inspection app. Refs.~\cite{dey2222, Dey2022, Online3} }
\end{figure} 


\begin{figure}[!tb]
	\begin{center}
		\begin{tabular}{c}
			\includegraphics[width=1.0\linewidth]{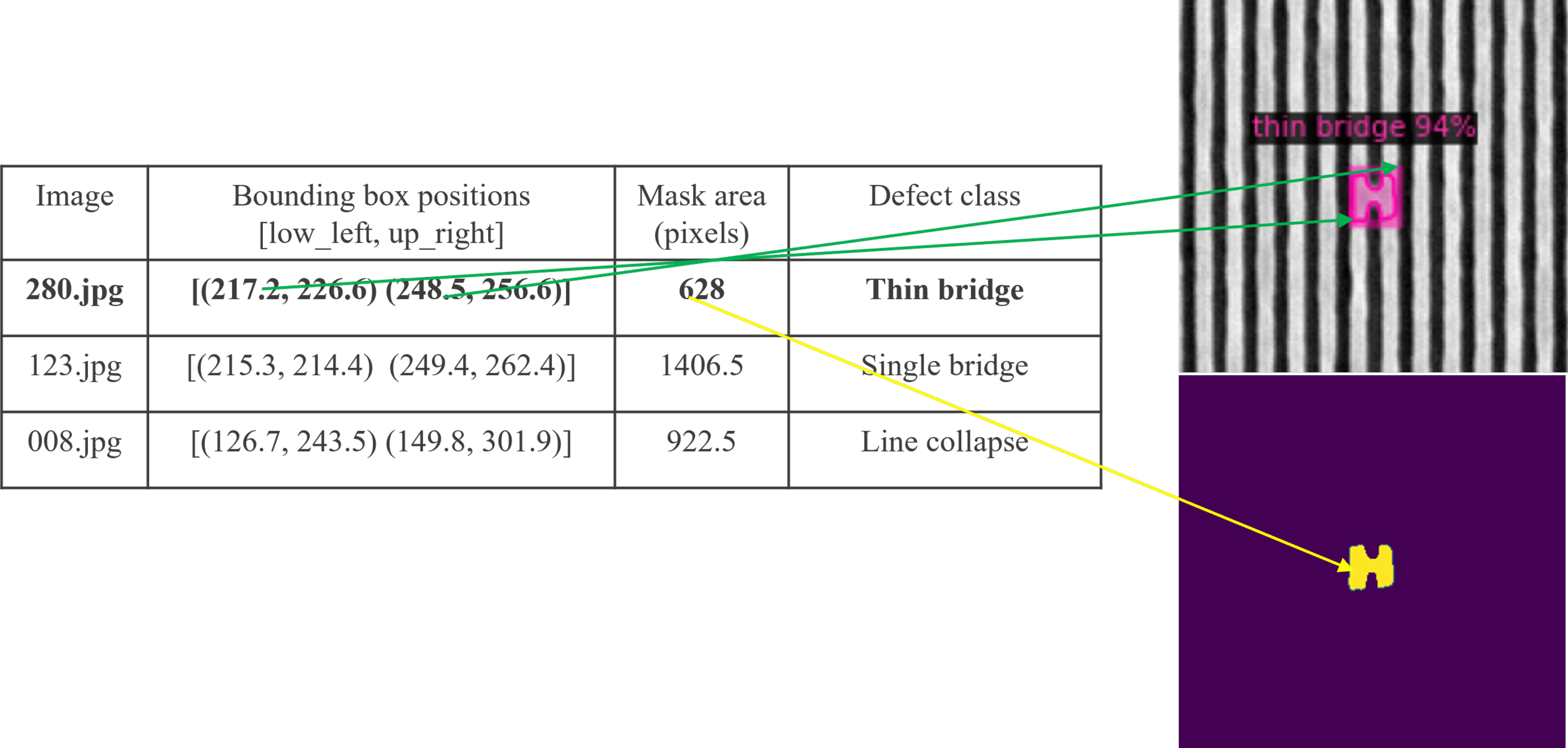}
		\end{tabular}
	\end{center}
	\caption 
	{ \label{fig:fig_7}
		Generated defect parameters by proposed ADCDS framework inference model. } 
\end{figure} 


\begin{figure}[!tb]
	\begin{center}
		\begin{tabular}{c}
			\includegraphics[width=0.80\linewidth]{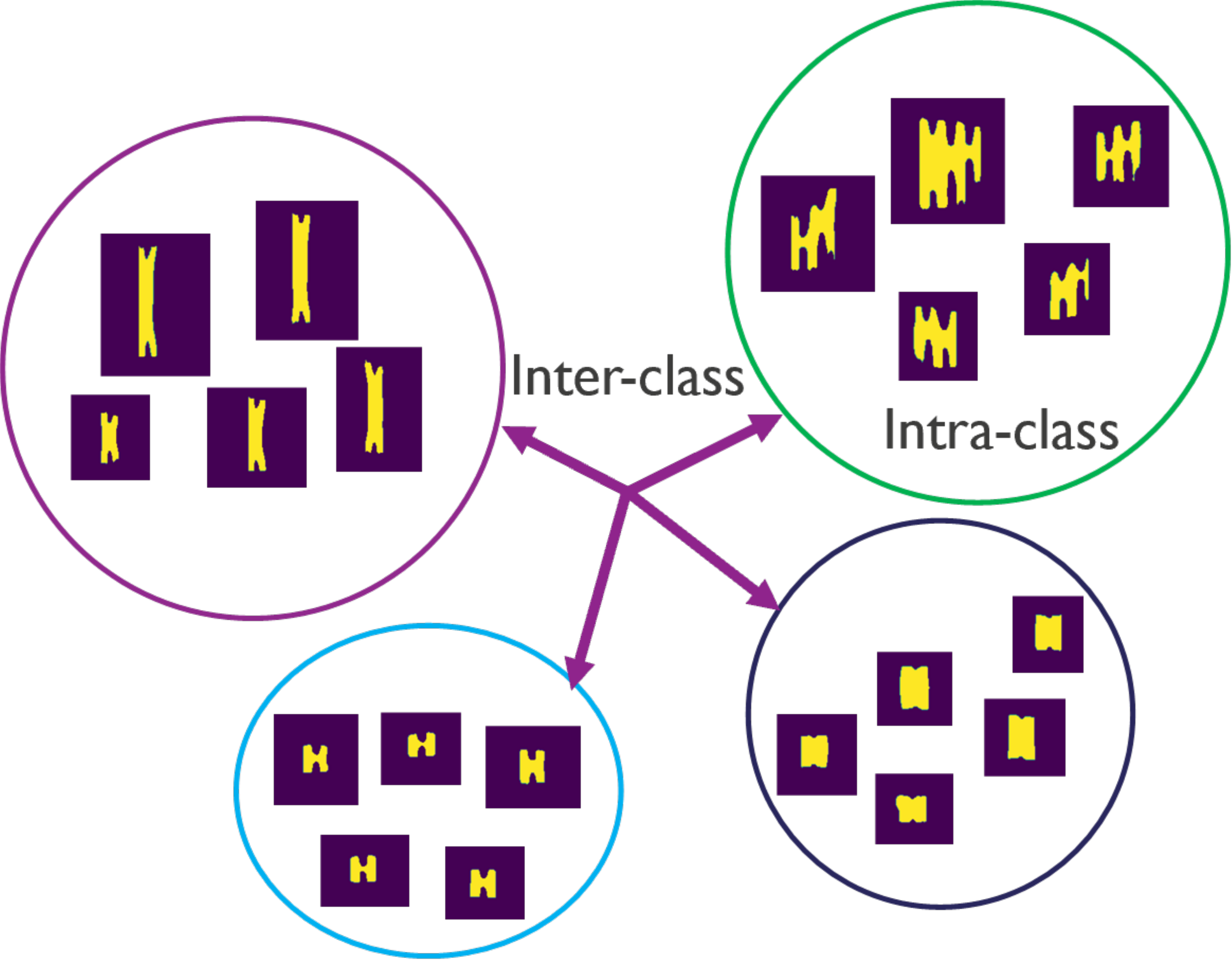}
		\end{tabular}
	\end{center}
	\caption 
	{ \label{fig:fig_8}
		Differentiation between different types of inter-class and intra-class stochastic defect patterns [based on surface area in terms of pixels]. } 
\end{figure} 


\section{Proposed Approach}
\label{section:Proposed Approach}

In this section, our proposed approach, based on Mask R-CNN, to detect different defects from SEM images, classify them according to their corresponding classes and segment the defect patterns based on their precise extent is presented. Our proposed ADCDS framework, which is illustrated in Fig.~\ref{fig:fig_2}, is trained, and evaluated using imec datasets (both Post-Litho and Post-Etch resist Wafer dataset) and classifies, detects, and segments the candidate defect types. To the best of our knowledge, this framework is the first to apply a novel robust supervised deep learning training scheme towards improved defect instance segmentation in SEM images with precise extent of defect as well as generating a mask for each defect category/instance. The proposed methodology is based on Mask R-CNN Ref.~\cite{he2017mask}, introduced by Facebook AI Research (FAIR), which combines the Faster R-CNN and FCN (Fully Connected Network) to get additional mask output other than the class and box outputs. Faster R-CNN Ref.~\cite{ren2015faster} is an object detection model consisting of two main components: a region proposal network (RPN) and a detection network. The RPN is a deep convolutional network that proposes regions of interest (ROI). The detection network focuses on these ROIs to detect objects of interest. The RPN shares convolutional features with the detection network to save computations. The model we used for segmenting SEM images is based on Mask R-CNN Ref.~\cite{he2017mask}. Mask R-CNN is a deep learning model that can perform object localization, semantic segmentation, and object segmentation for image data. Figure 3 provides an overview of the Mask R-CNN architecture. Mask R-CNN improves on the Faster R-CNN model Ref.~\cite{ren2015faster} by using an ROI-Align operation and by using separate classification and segmentation headers. Mask R-CNN replaces the ROI-Pooling operation applied in Faster R-CNN with an ROI-Align operation which allows for more precise instance segmentation. After quantization of the mapping process using convolutional layers, ROI-Pooling also quantizes feature maps during pooling which leads to data loss as some maps on the border of an ROI are not considered Ref.~\cite{Online}. ROI-Align performs interpolation between maps in the ROI in order to include information from all maps that fall (partially) within the ROI in the pooling process. This is especially important for instance segmentation which demands a more precise mask than bounding box-based object detection. Instance segmentation is handled by a convolutional network head that takes as input the ROIs and feature maps from the RPN.  This mask head works independently from a fully connected which classifies ROIs provided by the RPN as an object of interest or as background (no object of interest). Therefore, Mask R-CNN architecture adopts (1) ROI align operation, replacing ROI-pooling in Faster R-CNN, to allow to create precise instance segmentation masks. This is our goal here for extending previously proposed RetinaNet defect detector framework for defect classification and detection Refs.~\cite{dey2222, Dey2022, Online3}, (2) adds a network head as FCNN to produce the required defect instance segmentations, and (3) independent mask and detection header as shown in Fig.~\ref{fig:fig_3}. The proposed model is implemented using backbones as (a) faster+ResNet50+FPN, and (b) faster+ResNet101+FPN. Fig.~\ref{fig:fig_4} depicts the steps involved for training Mask R-CNN model for segmenting a defect instance and generating a binary mask for each ROI.

\section{Experiments}
\label{section:Experiments}

Our proposed defect detection framework is implemented using Detectron2 Ref.~\cite{wu2019detectron2} and PyTorch Ref.~\cite{NEURIPS2019-9015} library in the python programming environment. The Anaconda version was 4.9.2. Our model has been trained and evaluated on Lambda TensorBook with NVIDIA RTX 2080 MAX-Q GPU.

\subsection{Datasets}
The proposed Mask R-CNN based ADCDS framework [Classifier + Detector + Instance Segmentor] is trained and evaluated on both post-litho and post-etch P32 (Pitch 32 nm) resist wafer dataset. The dataset consists of a total of 600 raw SEM images of (480 ×480) pixels in TIFF format with stochastic defects such as thin and/or single bridge, line-collapse, line-breaks, multi-line-bridge (horizontal/non- horizontal) as well as clean images without any such defects. The representative defect class images from this dataset are already shown in Fig.~\ref{fig:fig1} (a) – (f). We have manually labelled 540 SEM images (480 training images,60 validation images) using Labelme Ref.~\cite{wada2018labelme} graphical image annotation tool. The defect labelling approach comprises varied defect representative and challenging condition instances and as per naming convention in Table~\ref{tab:Table1}. The dataset was divided into a training set, a validation set, and a test set as shown in Table~\ref{tab:Table2}. We have a total of 480 defect instances of these 6 different defect classes for training and 60 instances for validation. To comply with training criteria, we converted all images with “.tiff” format into “.jpg” format. We also implemented different data-augmentation techniques (as rotation, translation, shearing, scaling, flipping along X-axis and Y-axis, contrast, brightness, hue, and saturation) to balance/increase the diversity of training dataset defect patterns. We did not consider using any digital twins or synthetic datasets as cited in some previous citations as that does not solve the purpose of tackling real FAB originated stochastic defectivity scenario. We have also excluded any fabricated dataset patterned with intentionally placed or programmed defect types.

\begin{table}[ht]
	\caption{DEFECT CLASS LABELING CONVENTION} 
	\label{tab:Table1}
	\begin{center}       
		\begin{tabular}[6pt]{|m{7cm}|m{6cm}|}
			\hline
			\textbf{DEFECT CATEGORY}	& \textbf{LABELLED AS}  \\
			\hline
			
			THIN BRIDGE	& \textbf{thin$\_$bridge} \\
			\hline
			
			SINGLE BRIDGE &	\textbf{single$\_$bridge}  \\
			\hline
			
			LINE-COLLAPSE &	\textbf{line$\_$collapse}  \\
			\hline
			
			LINE-BREAK	& \textbf{line$\_$break}   \\
			\hline
			
			MULTI-LINE-BRIDGE-HORIZONTAL &	\textbf{multi$\_$bridge$\_$h}  \\
			\hline
			
			MULTI-LINE-BRIDGE-NON-HORIZONTAL &	\textbf{multi$\_$bridge$\_$nh}  \\
			\hline
			
		\end{tabular}
	\end{center}
\end{table} 


\begin{table}[ht]
	\caption{DATA DISTRIBUTION OF DEFECT SEM IMAGES} 
	\label{tab:Table2}
	\begin{center}       
		\begin{tabular}[6pt]{|m{4cm}|m{3cm}|m{3cm}|m{3cm}|}
			\hline
			Class Name	& Train (480 images) &	Val (60 images) &	Test (60 images) \\
			\hline
			
			thin$\_$bridge	& 80&	10&	10 \\
			\hline
			
			single$\_$bridge&	80&	10	&10 \\
			\hline
			
			line$\_$collapse&	80&	10&	10\\
			\hline
			
			line$\_$break&	80	&10&	10\\
			\hline
			
			multi$\_$bridge$\_$h	&80	&10	&10\\
			\hline
			
			multi$\_$bridge$\_$nh	&80	&10	&10\\
			\hline
			
			Total Instances&	480	&60&	60\\
			\hline
			
		\end{tabular}
	\end{center}
\end{table} 

\subsection{Evaluation criteria}
We have considered Intersection over Union (IoU) Ref.~\cite{rezatofighi2019generalized} between the ground truth bounding box and the predicted bounding box $\geq$ 0.5. The “defect detection confidence score” metric is taken as 0.8. The proposed ADCDS framework overall performance is evaluated against mAP as Mean Average Precision, where mAP is calculated using the weighted average of precisions among all defect classes. AP or average precision provides the detection precision for one specific defect class.

\subsection{Training}
We have first trained experimentally the different individual backbone architectures (ResNet50, ResNet101) on our SEM image dataset as discussed in previous section independently. For the proposed experiments, we have selected training parameters and hyperparameters as: 5000 epochs, batch-size of 1, initial learning rate at 0.00025, other default configuration as offered in Detectron2. Table~\ref{tab:Table3} presents validation detection accuracy per defect class of ResNet50 architecture as backbone. After successful completion of model training, the trained model was applied to previously unseen dataset to demonstrate defect detection performance, accuracy, and robustness.


\begin{table}[ht]
	\caption{VALIDATION ACCURACY OF TRAINED MASK R-CNN BASED ADCDS FRAMEWORK. IOU @ 0.50/AP50} 
	\label{tab:Table3}
	\begin{center}       
		\begin{tabular}[6pt]{|m{5cm}|m{4cm}|m{4cm}|}
			\hline
			\multirow{2}{*}{Class Name}	&\multicolumn{2}{c|}{Mask-RCNN}\\ \cline{2-3}
			
			& BBox AP	&Segmentation AP \\
			\hline
			
			line$\_$collapse&	0.822&	0.822 \\
			\hline
			
			single$\_$bridge&	1.00&	1.00\\
			\hline
			
			thin$\_$bridge	& 1.00	&1.00\\
			\hline
			
			multi$\_$bridge$\_$nh	& 0.833	&0.515 \\
			\hline
			
			\textbf{mAP}&	\textbf{0.914}	& \textbf{0.834}\\
			\hline
			
		\end{tabular}
	\end{center}
\end{table} 


\begin{table}[ht]
	\caption{OVERALL TEST ACCURACY OF TRAINED MASK R-CNN BASED ADCDS FRAMEWORK. IOU @ 0.50/AP50} 
	\label{tab:Table4}
	\begin{center}       
		\begin{tabular}[6pt]{|m{5cm}|m{4cm}|m{4cm}|}
			\hline
			\multirow{2}{*}{Class Name}	&\multicolumn{2}{c|}{Mask-RCNN} \\
			\cline{2-3}
			
				&BBox AP	&Segmentation AP \\
			\hline
			
		line$\_$collapse&	0.891&	0.891 \\
		\hline
		
		single$\_$bridge&	1.00&	1.00\\
		\hline
		
		thin$\_$bridge	&1.00	&1.00\\
		\hline
		
		multi$\_$bridge$\_$nh	&0.851&	0.851\\
		\hline
		
		\textbf{mAP} & 	\textbf{0.936}	& \textbf{0.936} \\
		\hline
			
		\end{tabular}
	\end{center}
\end{table} 

\section{Results}
\label{section:Results}
Examples of typical defect classification, detection and corresponding segmentation results are shown in Fig.~\ref{fig:fig5} (a)- (h) as single bridge, thin bridge, multi-line bridge (non-horizontal), and line-collapse, respectively. Our trained Mask-R-CNN model achieves the detection precision (BBox$\_$AP) of 89.1\% for line-collapse, approximately 100.0\% for both single and thin-bridge, 85.1\% for multi-bridge (non-horizontal) defectivity, respectively. Also, pixel segmentation precision (Segmentation$\_$AP) follows similar results. Table~\ref{tab:Table4} provides the evaluation results for defect detection accuracies obtained per defect class as well as mAP on test images for the Mask R-CNN approach with score-threshold 0.5, with AP50. There is further scope of improvement of AP/mAP metrices by considering ensembling of different deep feature extractor networks as backbones as well as addition of new defect categories and fine-tuning the network parameters/ hyperparameters. Fig.~\ref{fig:fig_6} illustrates our designed web-based defect inspection app. This UI (User-Interface) is built using Streamlit library Ref.~\cite{Online2} in python script to deploy our proposed model. The prototype version of the original proposed software interface is available on imec server for defect classification, which enables different partners/vendors to run the application on their local servers/workstations on their own tool data. Using appropriate graphical widgets as shown in Fig.~\ref{fig:fig_6} enables uploading a dataset of SEM/EDR/Review-SEM images (ADI/AEI) and finally segregating and saving those images along with generated defect pattern masks in different folders according to their defect categorical classes in local machines. The defect inspection app also generates a central CSV file, which includes different parameters (such as surface area in terms of pixels, length, width, corresponding defect class etc.) for better classification analysis and understanding the root cause of the defects, as shown in Fig.~\ref{fig:fig_7}. Finally, Fig.~\ref{fig:fig_8} demonstrates how our proposed approach helps differentiating between different types of inter-class as well as intra-class stochastic defect patterns (as thin/single/multi-line/horizontal/non-horizontal), based on precise extent of defect in a semi-supervised method.

\section{Conclusion}
\label{section:Conclusion}
In this paper, we have developed a novel robust supervised deep learning training scheme to accurately classify, localize as well as segment different defect types in SEM images with high degree of accuracy, based on Mask R-CNN. The proposed ADCDS framework enables to extract and calibrate each segmented mask and quantify the pixels that make up each mask, which in turn enables us to count each categorical defect instances as well as to calculate the surface area in terms of pixels. The goal of this work is to make an attempt to solve the issue of better classification by applying Deep Learning (DL)-based algorithms, aiming at detecting and segmenting different types of inter-class stochastic defect patterns such as bridge, break, and line collapse as well as to differentiate accurately between intra-class multi-categorical defect bridge scenarios (as thin/single/multi-line/horizontal/non-horizontal) for aggressive pitches as well as thin resists (High NA applications). Our proposed defect detection framework is noise-agnostic, scale-invariant, contrast-agnostic as well as Litho-step (ADI/AEI) invariant.
Our future approach is to extend this research towards (1) generate defect classes and locations, (2) generate parameters for each defect, (3) use data to model defect transfer from litho to etch, and finally (4) expand to other SEM applications (Logic/CH structures) as well as use other sets of images as TEM/AFM etc. Another possibility is (1) experimentation with (a) different other state-of-the-art deep feature extractor networks as backbones as well as (b) different recent one-stage/ two-stage/ensemble detector algorithms to further improve the overall mAP accuracy as well as individual defect class AP accuracy and inference speed, (2) addition of new defect categories and experiment with fine tuning of the network parameters to further improve the overall mAP metric.

\bibliography{report} 
\bibliographystyle{spiebib} 

\end{document}